%% file: ms.tex
\pdfoutput=1

\documentclass[sigconf]{acmart}

\renewcommand\footnotetextcopyrightpermission[1]{} 
\usepackage[ruled, linesnumbered]{Sty/algorithm2e}
\usepackage{amsmath}
\usepackage{amssymb}
\usepackage{amsthm}
\usepackage{bm}

\usepackage{tabularx}
\usepackage{Sty/bbm}
\usepackage{Sty/multirow}
\usepackage{Sty/subcaption}
\usepackage{Sty/mathtools}
\usepackage{Sty/graphicx}
\graphicspath{ {image/} }

\newtheorem{property}{Property}

\input{macros}

\def\BibTeX{{\rm B\kern-.05em{\sc i\kern-.025em b}\kern-.08emT\kern-.1667em\lower.7ex\hbox{E}\kern-.125emX}}
    
\copyrightyear{2019}
\acmYear{2019}
\setcopyright{acmlicensed}
\acmConference[]{}
\acmBooktitle{}
\acmPrice{}
\acmDOI{}
\acmISBN{}

\begin{document}

\newcolumntype{C}[1]{>{\centering\arraybackslash}p{#1}}

%
\title{Statistically Discriminative Sub-trajectory Mining}

%

\author{Vo Nguyen Le Duy}
\affiliation{%
  \institution{Nagoya Institute of Technology}
}
\email{duy.mllab.nit@gmail.com}


\author{Takuto Sakuma}
\affiliation{%
  \institution{Nagoya Institute of Technology}
}
\email{sakuma.takuto@nitech.ac.jp }


\author{Taiju Ishiyama}
\affiliation{%
  \institution{Nagoya Institute of Technology}
}
\email{ishiyama.t.mllab.nit@gmail.com}


\author{Hiroki Toda}
\affiliation{%
  \institution{Nagoya Institute of Technology}
}
\email{toda.h.mllab.nit@gmail.com}


\author{Kazuya Nishi}
\affiliation{%
  \institution{Nagoya Institute of Technology}
}
\email{nishi.k.mllab.nit@gmail.com}


\author{Masayuki Karasuyama}
\affiliation{%
  \institution{Nagoya Institute of Technology}
}
\email{karasuyama@nitech.ac.jp}


\author{Yuta Okubo}
\affiliation{%
  \institution{Sompo Japan Nipponkoa Insurance}
}
\email{yookubo21@sjnk.co.jp}

\author{Masayuki Sunaga}
\affiliation{%
  \institution{Sompo Japan Nipponkoa Insurance}
}
\email{msunaga1@sjnk.co.jp}


\author{Yasuo Tabei}
\affiliation{%
  \institution{RIKEN}
}
\email{yasuo.tabei@gmail.com}

\author{Ichiro Takeuchi}
\affiliation{%
  \institution{Nagoya Institute of Technology}
}
\email{takeuchi.ichiro@nitech.ac.jp}

%
\renewcommand{\shortauthors}{Duy, et al.}

%
\begin{abstract}
\input{abst}
\end{abstract}

%
%

%

%
\keywords{Trajectory mining; discriminative pattern mining; statistical testing; multiple testing}

%
\maketitle

\input{sec1}
\input{sec2}
\input{sec3}

\input{sec4}

\input{sec5}

%

\bibliographystyle{ACM-Reference-Format}
\bibliography{ms}

\end{document}

%% file: macros.tex
\newcommand{\cD}{{\mathcal D}}

\newcommand{\cG}{{\mathcal G}}

\newcommand{\cP}{{\mathcal P}}

\newcommand{\cT}{{\mathcal T}}

\newcommand{\cX}{{\mathcal X}}

\newcommand{\veps}{\varepsilon}


%% file: abst.tex
We study the problem of \emph{discriminative sub-trajectory mining}.
Given two groups of trajectories, the goal of this problem is to extract moving patterns in the form of sub-trajectories which are more similar to sub-trajectories of one group and less similar to those of the other.
We propose a new method called \emph{Statistically Discriminative Sub-trajectory Mining (SDSM)} for this problem.
An advantage of the SDSM method is that the statistical significance of extracted sub-trajectories are properly controlled in the sense that the probability of finding a false positive sub-trajectory is less than a specified significance threshold $\alpha$ (e.g., 0.05), which is indispensable when the method is used in scientific or social studies under noisy environment.
Finding such \emph{statistically} discriminative sub-trajectories from massive trajectory dataset is both computationally and statistically challenging.
In the SDSM method, we resolve the difficulties by introducing a tree representation among sub-trajectories and running an efficient permutation-based statistical inference method on the tree.
To the best of our knowledge, SDSM is the first method that can efficiently extract statistically discriminative sub-trajectories from massive trajectory dataset. 
We illustrate the effectiveness and scalability of the SDSM method by applying it to a real-world dataset with 1,000,000 trajectories which contains 16,723,602,505 sub-trajectories.

%% file: sec1.tex
\section{Introduction}

\begin{figure}[t]
\centering
\includegraphics[width=\linewidth]{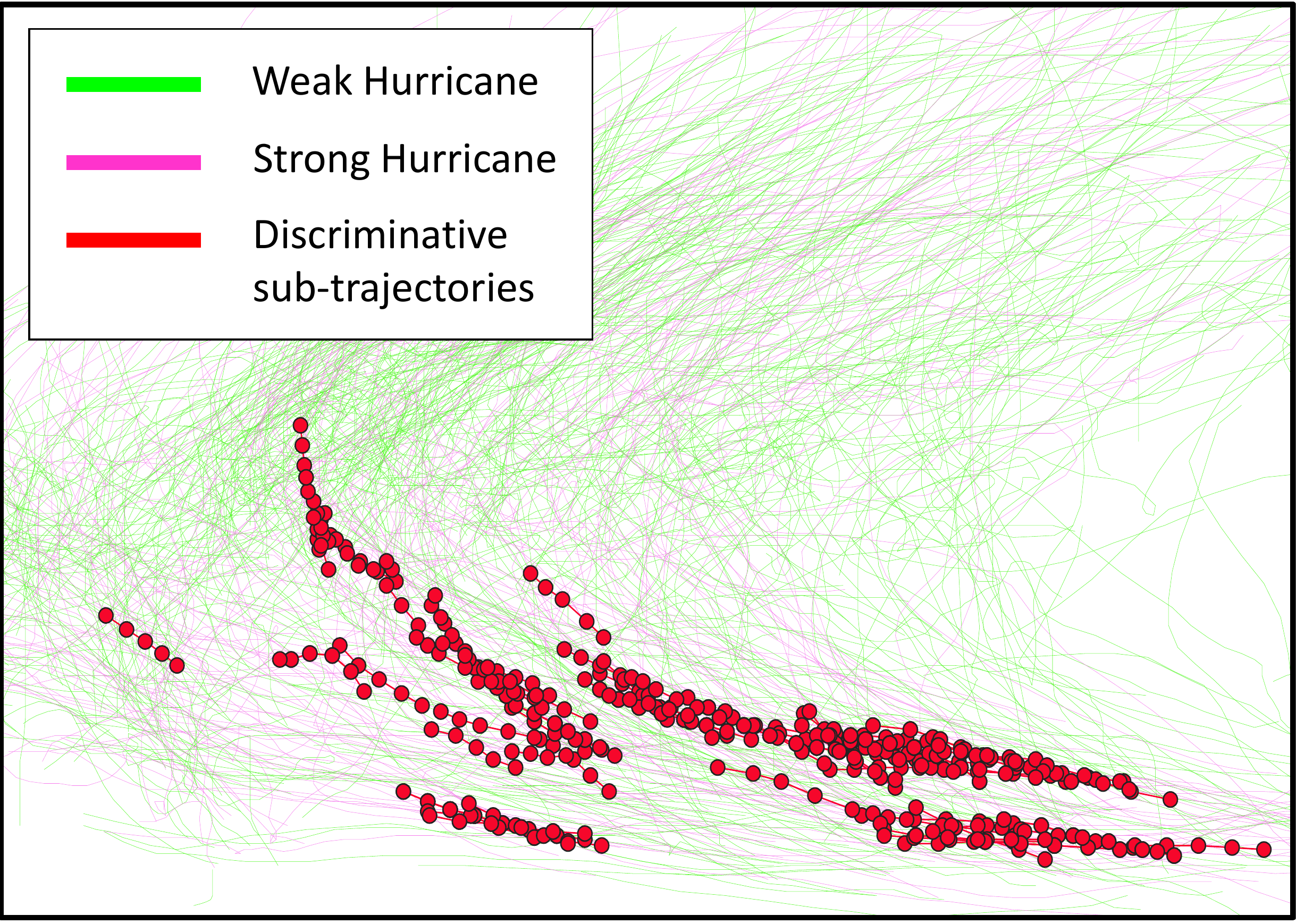}
\caption{Illustration of the discriminative sub-trajectories. Sub-trajectories (red) are determined to be discriminative sub-trajectories of strong hurricane group (pink) because they are similar to many sub-trajectories belonging to this group and satisfy statistical test criteria.}
\label{fig:ex_hurricane}
\end{figure}

Due to the rapid advance in location measurement technologies of moving objects, knowledge discovery from trajectory data, referred to as \textit{Trajectory Mining}, has been increasingly important \cite{zheng2015trajectory}. 
There is a variety of tasks in trajectory mining, e.g., 
discovering groups of objects that move together \cite{tang2012discovery, zheng2013discovery}, 
finding representative or common trajectories shared by different moving objects \cite{lee2007trajectory, li2010incremental, zheng2018incremental},
identifying groups of moving objects that travel a common sequence of locations \cite{wang2014travel}, and so on.

In this paper, we focus on the problem of \textit{discriminative sub-trajectory mining}. 
Given two groups of trajectories, the goal of discriminative sub-trajectory mining is to find moving patterns in the form of sub-trajectories which are more similar to sub-trajectories in one group and less similar to those in the other group. 
In many areas of scientific and social studies, discriminative sub-trajectory mining would be useful for finding moving patterns that are specific to certain experimental, environmental or social conditions. 
%

A naive solution for discriminative sub-trajectory mining is two-step approach, in which frequent/representative sub-trajectories are first extracted to construct a dictionary of sub-trajectories, and then discriminative sub-trajectories are selected from the dictionary. 
However, this naive approach is far less efficient than directly finding only discriminative sub-trajectories because we need to handle so many frequent/representative but non-discriminative common sub-trajectories in the first dictionary construction step.

In machine learning (ML) and statistics, the problem of finding discriminative features from a labeled dataset has been studied in the context of feature selection \cite{fan2008sure, pati1993orthogonal, tibshirani1996regression}. 
Among many possible criteria for discriminative features, we employ statistical \textit{\textit{p}-values} as the criteria of discriminative sub-trajectories. 
In scientific and social studies, \textit{p}-values are commonly used for quantifying the statistical significance of the findings because they enable us to properly control the false finding probability. 
Unfortunately however, when \textit{p}-values are computed for a large number of sub-trajectories, the \textit{multiple testing bias} \cite{shaffer1995multiple, dudoit2003multiple, benjamini1995controlling, benjamini2005false} must be properly corrected.
Multiple testing correction for a huge number of all possible sub-trajectories is both computationally and statistically challenging.

In this paper, we say that a sub-trajectory is \textit{statistically discriminative} if it satisfies the following two conditions.
\begin{enumerate}
	\item It is similar to many sub-trajectories of one group and different from (or similar to few) sub-trajectories of the other. 
	\item The properly adjusted \textit{p}-value for multiple testing correction is smaller than the predefined \textit{significance level} $\alpha$ (e.g., $0.05$).
\end{enumerate}

Figure \ref{fig:ex_hurricane} shows an example of discriminative sub-trajectories for hurricane dataset (see \S \ref{sec:experiments} for the details). 
This dataset is divided into two groups: weak hurricanes (green) and strong hurricanes (pink). 
The goal of this discriminative sub-trajectory mining task is to find statistically discriminative sub-trajectories that are highly associated with strong hurricane group (red).

To the best of our knowledge, there is no existing method that can efficiently find statistically discriminative sub-trajectories from massive trajectory data. 
Therefore, we propose a novel method, called SDSM (\underline{S}tatistically \underline{D}iscriminative \underline{S}ub-trajectory \underline{M}ining).
The main contributions of this paper are as follows:

\begin{enumerate}
	\item We introduce a statistical approach with multiple testing correction for assessing the significance of the discriminative ability of moving patterns in the form of sub-trajectories.
	\item We propose an algorithm for finding discriminative sub-trajectories by simultaneously performing sub-trajectory mining and multiple testing correction. The key idea is to take into account several properties between sub-trajectories.
	\item We conduct experiments on three real-world datasets. With the experiments, we offer the evidence that our algorithm (1) produces statistically significant results for analyzing moving objects, (2) can deal with a big dataset (e.g., 1,000,000 trajectories) and (3) works for different data from different fields. 
\end{enumerate}

\subsection{Related Works}
 Trajectory mining has been intensively studied for various tasks, such as trajectory pattern mining, trajectory clustering, and trajectory classification. 
Here, we restrict our attention to the related works on sub-trajectory mining, i.e, finding moving patterns in the form of sub-trajectories.
One of the most fundamental task of sub-trajectory mining is to discover frequent sub-trajectories \cite{lee2007trajectory, li2010incremental, zheng2018incremental}, in which the goal is to find representative moving patterns. 
Sub-trajectory mining is also useful for compressing a huge trajectory database. The goal here is to find a set of sub-trajectories which is used to approximately represent all the trajectories in the database \cite{zhao2018rest}. 
Given a trajectory dataset with group labels, we are often interested in constructing classifier, in which useful sub-trajectories for classification are selected by using feature selection methods \cite{ferrero2018movelets, lee2008traclass, patel2013incorporating}.
Although these sub-trajectories can be interpreted as discriminative, their statistical significances are not evaluated. 

\textit{Discriminative pattern mining} has been studied in standard pattern mining problems such as itemset mining and graph mining. 
Several different but related definitions of discriminative patterns have been proposed in the literature such as contrast sets \cite{bay2001detecting}, emerging patterns \cite{dong1999efficient} and subgroups \cite{klosgen1996explora, wrobel1997algorithm}. %
Although such patterns are used under different names, they are all interpreted as the methods for discriminative pattern discovery.
In some trajectory mining studies \cite{lee2011mining, sakuma2019efficient}, trajectories are first transformed into sequences of symbols, and sequence mining analogous to the above discriminative mining approaches are used for classifying two or more groups of trajectories. In this paper, we consider methods for directly handling spatial information and do not consider methods for symbolized trajectory data.

The problem of mining \textit{statistically significant patterns} has been recently receiving some attention in data mining community \cite{terada2013fast, llinares2015fast}. 
Performing statistical evaluation based on \textit{p}-values is necessary because it helps to control the probability of false positive findings, i.e., some discovered patterns in the observed data might not exist in the wider population.
When simultaneously conducting tests on thousands of patterns, \textit{p}-values must be properly adjusted to avoid multiple testing bias problem.
The most commonly-used multiple testing measure is \textit{Family-Wise Error Rate (FWER)}, which is defined as the probability of producing at least one false discovery.
In order to properly control the FWER under certain significance level $\alpha$ (e.g., 0.05), several multiple testing correction methods have been studied. 

The most commonly used method is Bonferroni correction \cite{tarone1990modified}, in which the adjusted significance level $\delta$ is obtained by $\alpha / M$, where $M$ is number of hypothesis tests. 
However, when the number of tests $M$ is large, $\delta$ will be very small, leading to too conservative correction with too many false negatives.  
As another approach, Westfall-Young method (WY) \cite{westfall1993resampling} is proposed to control FWER by calculating $\delta$ based on a null distribution estimated from thousands of randomly permuted datasets. 
The main limitation of WY method is that a large amount of computing time is required when the number of patterns to be considered is large. 
Recently, Fast Westfall-Young (FastWY) \cite{terada2013fast} and Westfall-Young Light \cite{llinares2015fast} have been proposed to accelerate the WY method in data mining community. 
Nevertheless, these methods are designed for itemset mining or graph mining tasks and can not be directly applied to trajectory data.

\subsection{Notations}
We use the following notations. For any natural number $n$, we define $[n] := \{1, ..., n\}$. 
The indicator function is written as $\mathbbm{1}[\cdot]$, i.e., $\mathbbm{1}[z] = 1$ if $z$ is true, and $\mathbbm{1}[z] = 0$	otherwise.


%% file: sec2.tex
\section{Problem Statement} \label{sec:problem_statement}

In this section, we first define several concepts for studying discriminative sub-trajectory mining in \S \ref{sub:definitions}. Then, we formulate the statistical test and multiple testing correction for discriminative sub-trajectories in \S \ref{sub:stats_dis_sub_traj}.


\subsection{Definitions} \label{sub:definitions}

\textbf{Raw trajectory.}
Let us consider a set of $n$ labeled trajectories denoted as 
$\bm \cD := \{(\cT_i, g_i)\}_{i \in [n]}$,
where
$\cT_i$
is the $i^{\rm th}$ trajectory,
and
$g_i \in \{\pm 1\}$
is the group label of $\cT_i$.
Here,
each trajectory
$\cT_i$
is represented by a sequence of time-ordered locations
$\cT_i := \{\bm p^t_i\}_{t \in [m_i]}$,
where
$m_i$
is length of the trajectory 
and
$\bm p^t_i$
is the vector representing the location at $t^{\rm th}$ timestamp.
For example, 
when we consider trajectories in 2D Euclidean space,
$\bm p^t_i$
is the vector representing x-coordinate and y-coordinate. 
We denote
a group of trajectories whose labels are positive
($g_i = +1$) as $\mathcal{G}_{+}$,
and
a group of trajectories whose labels are negative
($g_i = -1$)
as $\mathcal{G}_{-}$.
The sizes of each group are denoted as 
$n_{+} \coloneqq |\mathcal{G}_{+}|$
and
$n_{-} \coloneqq |\mathcal{G}_{-}|$,
respectively.
We call each $\cT_i, i \in [n]$, as \emph{raw trajectory} in contrast to \emph{sub-trajectory} defined in the next paragraph.  


\textbf{Sub-trajectory.}
In this paper, we consider sub-trajectories of $\cT_i$, $\forall i \in [n]$, whose length is greater than or equal to $L$, where $L$ is a tuning parameter. 
A sub-trajectory, represented as 
$T^{(s,e)}_i := \{\bm p^s_i, \ldots, \bm p^e_i\}$,
is 
a sequence of consecutive points of the raw trajectory
$\cT_i$
which starts from index $s$ and ends at index $e$,
where
$s$ and $e$
satisfy 
$1 \le s < e \le m_i$
and
$e - s + 1 \ge L$.
A notation
$T^{(s, e)}_i \sqsubseteq \cT_i$
indicates that 
$T^{(s, e)}_i$
is
a sub-trajectory
of a longer trajectory
$\cT_i$.
We denote a set of all possible sub-trajectories as 
$\bm T := \{T^{(s, e)}_i \mid 1 \le s < e \le m_i, e - s + 1\ge L, \forall i \in [n]\}$, and denote its size as $N := |\bm T|$.
We note that the number of all possible sub-trajectories $N$ is very large.


\textbf{Distance metric between sub-trajectories.}
In this work, it is important to define an appropriate distance metric between sub-trajectories. 
Here, we introduce a class of distance metrics called \emph{average-top-$K$-max} distance, which includes \emph{max} distance and \emph{average} distance as special cases. 
The distance between two different sub-trajectories
$T_i^{(s, s + \ell)} \text{ and } T_{i^\prime}^{(s^\prime, s^\prime + \ell)}$ with the same length
is defined based on
the \emph{pointwise distance}
$d(\bm p^{s + \tau}_i, \bm p^{s^\prime + \tau}_{i^\prime})$
for $\tau = 0, 1, ..., \ell$.
As the pointwise distance, we simply employ the Euclidean distance in this paper, but any other proper distance can be used instead.
The average-top-$K$-max distance between 
two different sub-trajectories
$T_i^{(s,s + \ell)}$
and
$T_{i^\prime}^{(s^\prime, s^\prime + \ell)}$ 
is defined as
\[
{\rm dist}_K(T_i^{(s,s + \ell)}, T_{i^\prime}^{(s^\prime, s^\prime + \ell)}) = \frac{1}{K} \sum_{k=1}^K d_{(k)} \quad \text{for } K \leq \ell, 
\] 
where
$d_{(k)}, k \in [K]$,
is the
$k^{\rm th}$ largest pointwise distance among the list 
$\{d(\bm p^{s + \tau}_i, \bm p^{s^\prime + \tau}_{i^\prime}) \}_{\tau \in \{0, 1, ..., \ell\}}$.
From the definition of average-top-$K$-max distance, the following property can be obviously derived:

\begin{property} \label{distance_property}
\textbf{Distance property.}
 The average-top-$K$-max distance satisfies
 \[
  {\rm dist}_K(T_i^{(s,e)}, T_{i^\prime}^{(s^\prime, e^\prime)}) \le {\rm dist}_K(T_i^{(s,e + \Delta l)}, T_{i^\prime}^{(s^\prime, e^\prime + \Delta l)}),
 \]
 where $e - s = e^\prime - s^\prime$ and $\Delta l \ge 0 $.
\end{property}


\textbf{Support and $\veps$-similar-neighborhood of sub-trajectory.}
We define \emph{$\veps$-similar-neighborhood} for each sub-trajectory $T_i^{(s,e)} \in \bm T$ as 
\[
N_\veps(T_i^{(s, e)}) := \{T_{i^\prime}^{(s^\prime, e^\prime)} \mid {\rm dist}_K(T_i^{(s, e)}, T_{i^\prime}^{(s^\prime, e^\prime)}) \le \veps\},
\]
where $\veps$ is a distance threshold and $e - s = e^\prime - s^\prime$ .
Then, we define the support of $T_i^{(s, e)}$ with respect to a subset of raw trajectories $\cG \subseteq [n]$ as 
\[
{\rm sup}_\cG(T_i^{(s, e)}) := |\{i^\prime \in \cG \mid \exists~T_{i^\prime}^{(s^\prime, e^\prime)} \sqsubseteq \cT_{i^\prime}, T_{i^\prime}^{(s^\prime, e^\prime)} \in N_\veps(T_i^{(s, e)})\}|,
\]
which indicates the number of raw trajectories in $\cG$ containing at least one sub-trajectory whose distance from the sub-trajectory $T_i^{(s, e)}$ is smaller than or equal to $\veps$.


\textbf{Discriminative sub-trajectory.}
Now, we are ready to define a discriminative sub-trajectory $T_i^{(s,e)}$ based on 
${\rm sup}_{\cG+}(T_i^{(s, e)})$
and 
${\rm sup}_{\cG-}(T_i^{(s, e)})$ 
for
$T_i^{(s,e)} \in \bm T$. 
To this end, we consider a contingency table as shown in Table \ref{tab:contingency_table},
where
``\#$\veps$-neighbors'' (resp. ``\#non-$\veps$-neighbors'') indicates the number of raw trajectories which contain (resp. do not contain) sub-trajectories whose distance from $T_i^{(s, e)}$ is smaller than $\veps$. 

Given the contingency table for each of the sub-trajectory $T_i^{(s, e)} \in \bm T$, we can quantify the statistical significance of the \emph{discriminative ability} of $T_i^{(s, e)}$ in the form of $p$-value.
Although there are several hypothesis testing methods for assessing the association between rows and columns in the contingency table, we employ Fisher's exact test \cite{fisher1922interpretation}. 

\begin{table}[t]
\caption{Contingency table for a sub-trajectory $T_i^{(s,e)}$.}
\label{tab:contingency_table}
\begin{center}
\begin{tabular}{l||c|c||c} 
& ~~~\#$\veps$-neighbors~~~ & \#non-$\veps$-neighbors & Total\\ 
\hline \hline
$g = +1$& ${\rm sup}_{\cG+}(T_i^{(s, e)})$ & $n_{+} - {\rm sup}_{\cG+}(T_i^{(s, e)})$ & $n_{+}$ \\ 
\hline
$g = -1$& ${\rm sup}_{\cG-}(T_i^{(s, e)})$ & $n_{-} - {\rm sup}_{\cG-}(T_i^{(s, e)})$ & $n_{-}$ \\ 
\hline \hline
Total & ${\rm sup}_{[n]}(T_i^{(s, e)})$ & $n - {\rm sup}_{[n]}(T_i^{(s, e)})$ & $n$\\ 
\end{tabular}
\end{center}
\end{table}


\subsection{Statistical Test for Discriminative Sub-trajectory} \label{sub:stats_dis_sub_traj}

\textbf{Fisher's exact test (FET).} To assess the discriminative ability of each sub-trajectory $T_i^{(s, e)} \in \bm T$, we need to determine whether the rows (group labels) and the columns (\#$\veps$-neighbors) in Table \ref{tab:contingency_table} are significantly associated or not. 
In order to quantify the statistical significance, we perform FET. 
In the null hypothesis of FET, it is assumed that the rows and the columns are statistically independent.
If we can find the information from data that provides evidence against the assumption of the null hypothesis, we can claim that the sub-trajectory $T_i^{(s,e)}$ is statistically discriminative. 
In FET, the marginal distribution of the $2\times2$ contingency table are fixed. 
Then, under the null hypothesis of the independence, the probability of observing ${\rm sup}_{\cG+}(T_i^{(s, e)}) = x$ in the upper-left part of the $2\times2$ contingency table follows the hypergeometric distribution, and is calculated as
\begin{align*}
	\mathcal{P}(x) = \frac{\displaystyle \binom{\ n_{+}\ }{ x }\ \displaystyle \binom{\ n_{-}\ }{\ {\rm sup}_{[n]}(T_i^{(s,e)}) - x\ } }{\displaystyle \binom{\ n\ }{\ {\rm sup}_{[n]}(T_i^{(s,e)})\ }}.
\end{align*}
The \textit{p}-value is defined as the probability of finding the observed, or more extreme, associations when the null hypothesis is true. 
The \textit{p}-value of sub-trajectory $T_i^{(s,e)}$ is given by the cumulative probability of all possible values that are at least as extreme as the one observed in the data, i.e., 
\begin{align*}
	p(T_i^{(s,e)}) = \sum \limits_{x \in \cX \mid \cP(x) \leq \cP( {\rm sup}_{\cG+}(T_i^{(s, e)}))} \cP(x),
\end{align*}
where $\cX := \{\max\{0, {\rm sup}_{[n]}(T_i^{(s,e)}) - n_{-}\}, ..., \min\{n_{+}, {\rm sup}_{[n]}(T_i^{(s,e)})\}\}$.
The smaller the \textit{p}-value is, the higher the probability of rejecting the null hypothesis is. 
In other words, the small \textit{p}-value indicates the high probability of the sub-trajectory to be discriminative. Thus, it is reasonable to find a set of sub-trajectories whose \textit{p}-values are sufficiently small.

\textbf{Family-wise error rate (FWER) control for multiple testing corrections.}
The advantage of using \textit{p}-value as a criterion is that the probability of false positive findings can be properly managed, which is important for controlling the quality of scientific or social findings. 
Unfortunately, when \textit{p}-values are computed for a large number of sub-trajectories, the problem of multiple testing bias arises. 
Specifically, if we select sub-trajectories whose \textit{p}-values are smaller than a certain threshold $\alpha$ (e.g., 0.05), then the probability of finding at least one false-positive sub-trajectory is far greater than the specified $\alpha$. 
For correcting the multiple testing bias, we control FWER, which is the probability of finding at least one false positive, to be smaller than $\alpha$. 
To this end, we can only select sub-trajectories whose \textit{p}-values are smaller than an \textit{adjusted significance level} $\delta$, which is usually much smaller than $\alpha$. 
The FWER obtained by an adjusted significance level $\delta$ is denoted as FWER($\delta$), and the largest $\delta$ whose FWER($\delta$) $\leq \alpha$ is denoted as $\delta^{\ast} = \max\{\delta\ |\ {\rm FWER}(\delta) \leq \alpha\}.$

We use Westfall-Young (WY) method for controlling FWER. 
Specifically, $B$ (e.g., 1000) randomized datasets are generated by randomly permuting the labels $\{g_i\}_{i \in [n]}$. 
Then, the FWER($\delta$) is estimated as ${\rm FWER}(\delta) = \frac{1}{B}\sum_{b=1}^B \mathbbm{1}[p_{\rm min}^{(b)} \leq \delta]$,
where $p_{\rm min}^{(b)}$ is the smallest \textit{p}-value in the $b^{\rm th}$ permuted dataset.
Then, $\alpha$-quantile of the minimum \textit{p}-value distribution is used to estimate the optimal adjusted significance level $\delta^{\ast}$. 
If we enumerate all the sub-trajectories whose \textit{p}-values are less than the estimated $\delta^{\ast}$, the probability of finding at least one false positive discriminative sub-trajectory would be smaller than $\alpha$.
In other words, we can consider a sub-trajectory is discriminative when its adjusted \textit{p}-value, 
defined as $p_{\rm adj}(T_i^{(s,e)}) = p(T_i^{(s,e)}) \times (\alpha/\delta^{\ast})$, is smaller than $\alpha$.

In this paper, we employ WY method for finding discriminative sub-trajectories whose FWER is properly controlled. 
However, the computational cost of applying WY method to our problem is extremely large, since it requires us to compute all the $N$ \textit{p}-values for all possible sub-trajectories $B = 1000$ times. 
In the next section, we will introduce a new method, called SDSM, for resolving this computational challenge, and demonstrate in \S \ref{sec:experiments} that the SDSM method can be successfully applied to a trajectory dataset with $n = $ 1,000,000, in which the number of all possible sub-trajectories is $N = $ 16,723,602,505.

%% file: sec3.tex

\section{Statistically Discriminative Sub-trajectory Mining (SDSM)} \label{sec:proposed_method}

In this section, we present the SDSM method as our main contribution. 
Our goal is to develop an algorithm which can efficiently enumerate all the sub-trajectories whose estimated FWER by WY method is smaller than $\alpha$. 
To achieve the goal, we need to resolve the following two challenges. 
First, since the number of all possible sub-trajectories $N$ is too large to handle, we need to introduce a strategy being able to screen out majority of sub-trajectories whose FWER cannot be sufficiently small. 
Second, since the optimal adjusted significance level $\delta^{\ast}$ is unknown, we need to estimate it. 
To tackle these two challenges, we employ a tree representation of sub-trajectories, and develop a tree-pruning strategy. 
In \S \ref{sec:proposed_method:tree_representation}, we discuss several properties between sub-trajectories which can be taken into account for constructing the tree representation. 
In \S \ref{sec:proposed_method:pruning_strategy}, we present the pruning strategy for efficiently screening out sub-trajectories which are irrelevant to FWER computation.
The pseudocode of the algorithm and detail explanation are presented in \S \ref{sec:proposed_method:algorithm}.


\subsection{Exploiting Properties between Sub-trajectories for Tree Representation} \label{sec:proposed_method:tree_representation}

We start by introducing several properties for constructing a tree representation of sub-trajectories.

\begin{figure}[t]
\centering
\includegraphics[width=0.9\linewidth]{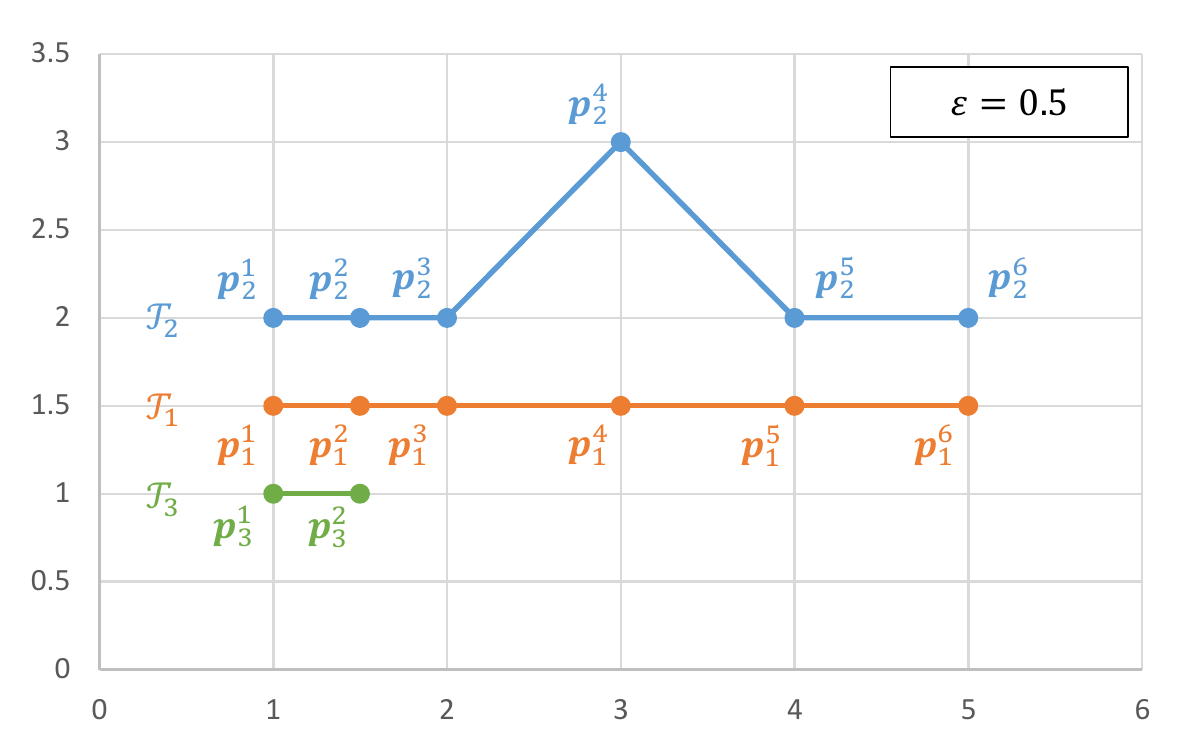}
\caption{Illustration of the properties between longer sub-trajectory and its sub-sequences in case of $n=3$, $K=2$  and $\veps = 0.5$. 
In this illustrative example, 
$N_\veps(T_1^{(1,2)}) = \{T_2^{(1,2)}, T_3^{(1,2)}\}$ because $dist_{K=2} (T_1^{(1,2)}, T_2^{(1,2)}) = \frac{1}{2}(0.5 + 0.5) = 0.5 \le \veps$ and  $dist_{K=2} (T_1^{(1,2)}, T_3^{(1,2)}) = 0.5 \le \veps$,
$N_\veps(T_1^{(1,3)}) = \{T_2^{(1,3)}\}$,
$N_\veps(T_1^{(1,4)}) = \emptyset$ because $dist_{K=2}(T_1^{(1,4)}, T_2^{(1,4)}) = \frac{1}{2} (0.5 + 1.5) = 1 > \veps$. 
It is obvious that when increasing the length of sub-trajectory, the cardinality of $\veps$-similar-neighborhood decreases, i.e.,$|N_\veps(T_1^{(1,2)})| > |N_\veps(T_1^{(1,3)})| > |N_\veps(T_1^{(1,4)})|$, and support decreases , i.e., ${\rm sup}_{[n]}(T_1^{(1,2)}) = 2 > {\rm sup}_{[n]}(T_1^{(1,3)}) = 1 > {\rm sup}_{[n]}(T_1^{(1,4)}) = 0$.
}
\label{fig:ex_property}
\end{figure}

\begin{property} \label{neighbor_property}
\textbf{$\veps$-similar-neighborhood property.} The cardinality of $\veps$-similar-neighborhood of sub-trajectory decreases when the length increases, i.e.,
\begin{align*}
	|N_\veps(T_i^{(s, e)})| \ge |N_\veps(T_i^{(s, e + \Delta l)})|, \ \Delta l \ge 0.
\end{align*}
\end{property}

Property \ref{neighbor_property} can be directly proved by the distance property (Property \ref{distance_property}).
Next, we describe the property between supports of longer and shorter sub-trajectories.

\begin{property} \label{support_property}
\textbf{Support property.} For any subset $\cG \subseteq [n]$, support of sub-trajectory decreases when the length increases, i.e.,
\begin{align*}
	{\rm sup}_{\cG}(T_i^{(s, e)}) \ge {\rm sup}_{\cG}(T_i^{(s, e + \Delta l)}), \ \Delta l \ge 0.
\end{align*}
\end{property}
Property \ref{support_property} is a direct consequence of Property \ref{neighbor_property}. 
Figure \ref{fig:ex_property} illustrates Properties \ref{neighbor_property} and \ref{support_property}.

We next define the relation between support and lower bound of \textit{p}-value of a sub-trajectory. 
Terada et al. \cite{terada2013fast} introduced the lower bound of \textit{p}-value and its monotonicity property for itemsets. 
We extend the concept for trajectory data. 
Let $L(T_i^{(s,e)})$ denote the lower bound of the \textit{p}-value of sub-trajectory $T_i^{(s,e)}$.
Since $L(T_i^{(s,e)})$ only depends on support ${\rm sup}_{[n]}(T_i^{(s,e)})$ and is independent from ${\rm sup}_{\cG_{+}}(T_i^{(s,e)})$ in the contingency table in Table \ref{tab:contingency_table}, the random permutation in WY method does not affect the lower bound.

\begin{property} \label{lower_bound}
\textbf{Lower bound of \textit{p}-value.} Given a sub-trajectory $T_i^{(s,e)}$, the lower bound of \textit{p}-value of $T_i^{(s,e)}$ in two-sided Fisher's exact test is computed as follows (we denote $L(T) := L(T_i^{(s,e)})$ for simplicity):
\begin{align*}
	L(T) &= {\rm min}(L_U(T), L_L(T),\\
	L_U(T) &= 
	\begin{cases}
		l({\rm sup}_{[n]}(T), {\rm sup}_{[n]}(T)), & \text{if}\ {\rm sup}_{[n]}(T) \leq n_{+},\\
		l(n_{+}, n_{+}), & {\rm otherwise},
	\end{cases} \\
	L_L(T) &= 
	\begin{cases}
		l({\rm sup}_{[n]}(T), 0), & \text{if}\ {\rm sup}_{[n]}(T) \leq n_{-},\\
		l(n_{-}, 0), & {\rm otherwise},
	\end{cases} \\
	\text{ \rm where }l(a,b) &= \frac{\displaystyle\binom{\ n_{+}\ }{b} \displaystyle\binom{\ n_{-}\ }{\ a - b\ }}{\displaystyle\binom{\ n\ }{\ a\ }}.
\end{align*}

\end{property}

\begin{property} \label{lower_bound_property}
\textbf{Monotonicity of lower bound.} The lower bound of sub-trajectory increases when the length increases, i.e.,
\begin{align*}
	L(T_i^{(s, e)}) \le L(T_i^{(s, e + \Delta l)}), \ \Delta l \ge 0.
\end{align*}
\end{property}
\begin{proof}
The lower bound monotonically increases as the support decreases \cite{terada2013fast}. 
Since the support of a longer sub-trajectory is always less than or equal to the support of its sub-sequences (Property \ref{support_property}), Property \ref{lower_bound_property} holds.
\end{proof}

\subsection{Pruning Strategy} \label{sec:proposed_method:pruning_strategy}

\begin{figure}[t]
\centering
\includegraphics[width=\linewidth]{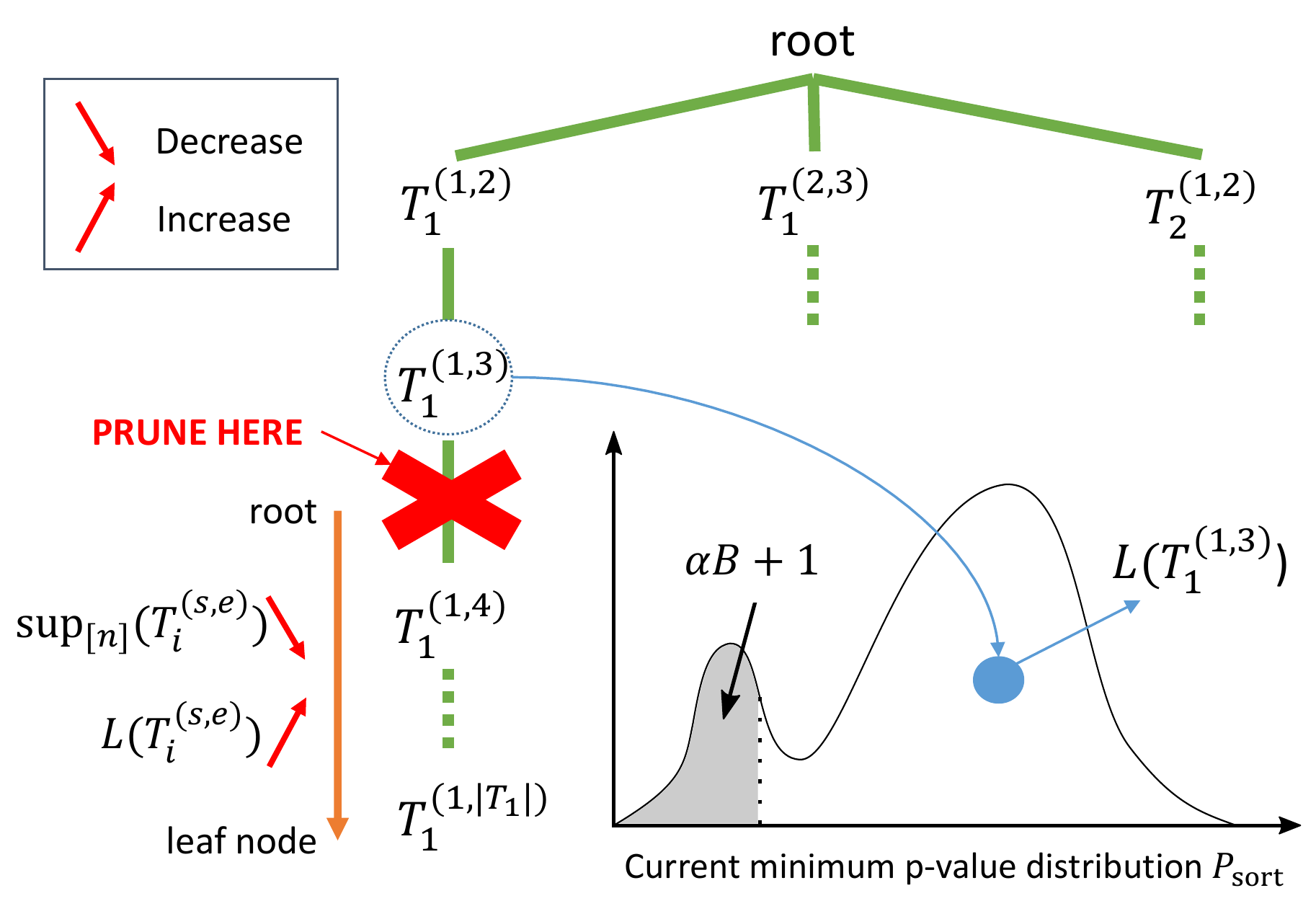}
\caption{Illustration of the pruning property. 
$B$ is the number of permutated datasets. $L(T_1^{(1,3)})$ indicates the lower bound of sub-trajectory $T_1^{(1,3)}$. 
From the root to leaf node, the support decreases and lower bound increases. 
Because $L(T_1^{(1,3)})$ is larger than ($\alpha B + 1$) smallest values of the current minimum \textit{p}-value distribution, $T_1^{(1,3)}$ and all the longer sub-trajectories must have \textit{p}-values larger than ($\alpha B + 1$) smallest values of the current null distribution from all the permuted datasets. 
Therefore, they do not affect the computation of $\delta^{\ast}$.}
\label{fig:pruning_property}
\end{figure}

Based on Properties \ref{lower_bound} and \ref{lower_bound_property}, we present a pruning property and a strategy to estimate the optimal adjusted significance level $\delta^{\ast}$. 
We only focus on calculating minimum \textit{p}-values which affect the computation of $\delta^{\ast}$ instead of precisely identifying the minimum \textit{p}-value distribution which requires minimum \textit{p}-values computations for all the permuted datasets.
To this end, we make two changes from the original WY method.

First, we change the formula to estimate the optimal adjusted significance level $\delta^{\ast}$ as 
\begin{align*}
	\delta^{\ast} = \max\{ P_{\rm sort}^{(b)}\ |\  P_{\rm sort}^{(b)} < P_{\rm sort}^{(\alpha B  + 1)}\},\ b \in [B],
\end{align*}
where $P_{\rm sort}$ is the sorted list of all minimum \textit{p}-values in the ascending order, $\alpha$ is the target significance level and $B$ is the number of permutations. 
From this formula, it is obvious that we only need to calculate $\lceil \alpha B  + 1 \rceil$ smallest minimum \textit{p}-values of the null distribution.

Second, we change the method so that the minimum \textit{p}-values of all the permuted datasets are simultaneously updated for each sub-trajectory.
Since we represent all the sub-trajectories as a tree, this process is performed in depth-first search manner. 
The main purpose of this change is that we can compute $\lceil \alpha B  + 1 \rceil$ smallest minimum \textit{p}-values without precisely calculating $\lfloor B - (\alpha B  + 1) \rfloor$ largest values by applying the pruning strategy. 
Next, we describe a useful property for the pruning process.

\begin{property}  \label{pruning_property}
\textbf{Property for pruning algorithm}.
Let $T_i^{(s,e)}$ be a current sub-trajectory that needs to be tested. If $L(T_i^{(s,e)}) \ge P_{\rm sort}^{(\alpha B + 1)}$ holds, no sub-trajectories $T_i^{(s, e + \Delta l)}$ that satisfies $T_i^{(s,e + \Delta l)} \sqsupseteq T_i^{(s,e)}$ affects the computation of $\delta^{\ast}$.
\end{property}

\begin{proof}
Since the permutation does not affect the lower bound (Property \ref{lower_bound}), and the monotonicity of lower bound  (Property \ref{lower_bound_property}), then we have:
\begin{align*}
	p^{(b)}(T_i^{(s,e + \Delta l)}) \ge L(T_i^{(s,e + \Delta l)}) \ge L(T_i^{(s,e)}),\ \Delta l \ge 0,
\end{align*}
where $p^{(b)}(T_i^{(s,e + \Delta l)})$ is the \textit{p}-value of $T_i^{(s,e + \Delta l)}$ at the $b^{\rm th}$ permuted dataset. This suggests that:
\begin{align*}
	L(T_i^{(s,e)}) \ge P_{\rm sort}^{(\alpha B + 1)} \Rightarrow p^{(b)}(T_i^{(s,e + \Delta l)})  \ge P_{\rm sort}^{(\alpha B + 1)},\ \Delta l \ge 0.
\end{align*}
Hence, Property \ref{pruning_property} holds.
\end{proof}

Property \ref{pruning_property} is illustrated in Figure \ref{fig:pruning_property}. 
In this figure, $P_{\rm sort}$ indicates the current minimum \textit{p}-value distribution. 
Because $L(T_1^{(1,3)}) > P_{\rm sort}^{(\alpha B + 1)}$, $p^{(b)}(T_1^{(1,3 + \Delta l)}) > P_{\rm sort}^{(\alpha B + 1)}$ for any $T_1^{(1, 3 + \Delta l)} \sqsupseteq T_1^{(1,3)}$ and $b \in [B]$.

Using this property, the following procedure can extract sub-trajectories and produce $\lceil \alpha B  + 1 \rceil$ smallest minimum \textit{p}-values simultaneously:

\begin{enumerate}
	\item Generate all the permuted datasets and set the initial $p_{\rm min}^{(b)}$ to $\alpha$ for each $b^{\rm th}$ permutation.
	\item For each branch of the tree, by starting at the shortest sub-trajectory as the first node, simultaneously update the minimum \textit{p}-value distribution and recursively explore the descendant nodes until all the nodes of the current branch are explored or the pruning condition is satisfied (Property \ref{pruning_property}).
	\item Output $\lceil \alpha B  + 1 \rceil$ smallest values of minimum \textit{p}-value distribution.
\end{enumerate}



\subsection{The Algorithm} \label{sec:proposed_method:algorithm}
The pseudocode of the proposed method SDSM is shown in Algorithm \ref{alg:algorithm}. 
It consists of two functions: (1) the \textit{Main} function and (2) the \textit{ProcessNext} function. 
In the \textit{Main} function, we present the initialization step and three main steps of our algorithm. 
In step 1: sub-trajectories are extracted and $\lceil \alpha B  + 1 \rceil$ smallest minimum \textit{p}-values of the null distribution are computed simultaneously. 
In step 2: optimal adjusted significance threshold $\delta^{\ast}$ is calculated. 
In step 3: the list of sub-trajectories whose \textit{p}-values are less than $\delta^{\ast}$ is constructed. 
\textit{ProcessNext} function describes the detail of the first main step.

{\SetAlgoNoLine
\begin{algorithm}

\SetKwProg{Function}{function}{}{}
\SetKwProg{Procedure}{procedure}{}{}

\let\oldnl\nl
\newcommand{\nonl}{\renewcommand{\nl}{\let\nl\oldnl}}

\DontPrintSemicolon 
\KwIn{\ Trajectory dataset $\bm \cT$, group labels $\textbf{g}$, distance threshold $\veps$, minimum length $L$, top-K-max $K$, number of permutations $B$ and target significance level $\alpha$.}
\KwOut{\ Enumerate discriminative sub-trajectories.}

\Procedure{Main \rm()}{
	\nonl // Initialization\;
	\For{$b \leftarrow 1$ {\rm to} $B$}{ \label{alg:start_initialization}
		$\textbf{g}^{(b)} \leftarrow$ permute($\textbf{g}$)\;
		$p_{\rm min}^{(b)} \leftarrow \alpha$\; 
	}\label{alg:end_initialization}
	\nonl // Extract sub-trajectory and Estimate null dustribution\;	
	\For{{\rm each} $\cT_i \in \bm \cT$}{ \label{alg:start_step1}
		\For{{\rm each length-}$L$ {\rm sub-trajectory} $T_i^{(s,e)} \sqsubseteq \cT_i$}{
			Compute $N_{\veps}(T_i^{(s,e)})$\; \label{calculate_neighbor}
			\textit{ProcessNext {\rm (}$T_i^{(s,e)}$, $N_{\veps}(T_i^{(s,e)})${\rm )}} 
		}
	}\label{alg:end_step1}

	\nonl // Calculate threshold $\delta^{\ast}$\;
	$P_{\rm sort} \leftarrow {\rm sort}(\{p_{\rm min}^{(b)}\}_{b=1}^B)$\; \label{alg:start_calibrate}
	$\delta^{\ast} \leftarrow \max(P_{\rm sort}^{(x)}\ |\ P_{\rm sort}^{(x)} < P_{\rm sort}^{(\alpha B+1)})$ \; \label{alg:end_calibrate}
	
	\nonl // Enumerate discriminative sub-trajectories\;
	Output sub-trajectories whose \textit{p}-values $ < \delta^{\ast}$ \label{alg:enumerate_discriminative}
}
\BlankLine
\Function{ProcessNext {\rm (}$T_i^{(s,e)}$, $N_{\veps}(T_i^{(s,e)})${\rm )}}{
	$P_{\rm sort} \leftarrow {\rm sort}(\{p_{\rm min}^{(b)}\}_{b=1}^B)$\; \label{alg:sort_list_p}
	Compute $L(T_i^{(s,e)})$\; \label{alg:compute_lower_bound}

	\If{$L(T_i^{(s,e)}) \ge P_{\rm sort}^{(\alpha B+1)}$}{ \label{alg:start_pruning}
		\Return\;
	}\label{alg:end_pruning}

	\For{$b \leftarrow 1$ {\rm to} $B$}{ \label{alg:start_update_min_p}
		\If{$L(T_i^{(s,e)}) < p_{\rm min}^{(b)}$}{
			$ p_{\rm min}^{(b)} \leftarrow \min\{\ p_{\rm min}^{(b)}, p^{(b)}(T_i^{(s,e)})\ \}$
		}
	}  \label{alg:end_update_min_p}

	\For{{\rm each} $T_{i^\prime}^{(s^\prime, e^\prime)} \in N_{\veps}(T_i^{(s,e)})$}{ \label{alg:start_continue_extraction}
		$d \leftarrow {\rm dist}_{K}(T_{i^\prime}^{(s^\prime, e^\prime + 1)}, T_i^{(s, e + 1)})$\;
		\If{$d \leq \veps$}{
			Add $T_{i^\prime}^{(s^\prime, e^\prime + 1)}$ into $N_{\veps}( T_i^{(s, e + 1)})$\;
		}
	} 
	\textit{ProcessNext{\rm (}$T_i^{(s, e + 1)}$, $N_{\veps}(T_i^{(s, e + 1)})${\rm )}} \label{alg:end_continue_extraction}
}

\caption{Statistically Discriminative Sub-trajectory Mining (SDSM)}
\label{alg:algorithm}
\end{algorithm}}

\subsubsection{The Main function \nopunct} \hfill\\
We start the algorithm with initialization step in Lines \ref{alg:start_initialization}-\ref{alg:end_initialization}. 
We precompute the permuted group labels for all the permuted datasets. 
Since we want to control FWER to be under $\alpha$, we do not need to consider sub-trajectories whose \textit{p}-values are greater than $\alpha$. 
The minimum \textit{p}-value $p_{\rm min}^{(b)}$ of each $b^{\rm th}$ permuted dataset is initialized at $\alpha$. 
Next, in Lines \ref{alg:start_step1}-\ref{alg:end_step1}, we perform the process of simultaneously extracting sub-trajectories and estimating null distribution. 
Sub-trajectories are extracted as nodes of the tree in which the child node $T_i^{(s,e + \Delta l)}$ indicates the longer sub-trajectory of the parent node $T_i^{(s,e)}$ and ${\rm sup}_{[n]}(T_i^{(s,e + \Delta l)}) \leq {\rm sup}_{[n]}(T_i^{(s,e)})$ holds. 
The first node of each branch of the tree is a sub-trajectory with the predefined minimum length $L$. 
Next, in Line \ref{calculate_neighbor}, we find the $\veps$-similar-neighborhood $N_{\veps}(T_i^{(s,e)})$ because it is used to compute the support of $T_i^{(s,e)}$ and generate $\veps$-similar-neighborhood of longer sequences. 
After the first node of the branch is initialized, the \textit{ProcessNext} function is called to simultaneously explore the current branch and update the null distribution. 

\subsubsection{The ProcessNext function \nopunct}\hfill\\
The \textit{ProcessNext} function processes one sub-trajectory at a time. 
First, in Line \ref{alg:sort_list_p}, we sort the current list of minimum \textit{p}-values in the ascending order and set them to $P_{\rm sort}$. 
We then compute the lower bound of current sub-trajectory (current node) in Line \ref{alg:compute_lower_bound}. 
Next, between Lines \ref{alg:start_pruning} and \ref{alg:end_pruning}, we check the pruning condition. 
If the lower bound of current sub-trajectory $L(T_i^{(s,e)}) \ge P_{\rm sort}^{(\alpha B + 1)}$, we stop the exploration process of the current branch. 

If the pruning condition is not satisfied, Lines \ref{alg:start_update_min_p}-\ref{alg:end_update_min_p} continue to be processed to update the minimum \textit{p}-value distribution. 
For all permutations $b \in [B]$, if the lower bound $L(T_i^{(s,e)}) < p_{\rm min}^{(b)}$, we then update $p_{\rm min}^{(b)}$ of $b^{\rm th}$ permutation if $p^{(b)}(T_i^{(s,e)}) < p_{\rm min}^{(b)}$. 
%

Finally, in Lines \ref{alg:start_continue_extraction}-\ref{alg:end_continue_extraction}, the child node (longer sub-trajectory) is continued to be explored and the process of updating null distribution is performed on each new child node by recalling the \textit{ProcessNext} function. 
The $\veps$-similar-neighborhood $N_{\veps}(T_i^{(s, e + 1)})$ of longer sub-trajectory $T_i^{(s, e + 1)}$ can be easily derived based on the  $N_{\veps}(T_i^{(s,e)})$ of the current sub-trajectory $T_i^{(s,e)}$.

This function simultaneously performs the sub-trajectory extraction and null distribution updating process until all the $\lceil \alpha B  + 1 \rceil$ smallest minimum \textit{p}-values are completely calculated. 
Then, the optimal adjusted significance threshold $\delta^{\ast}$  is calibrated in Lines \ref{alg:start_calibrate}-\ref{alg:end_calibrate} and discriminative sub-trajectories are outputed in Line \ref{alg:enumerate_discriminative}.



%% file: sec4.tex

\section{Experimental Evaluation} \label{sec:experiments}
In this section, we evaluate the effectiveness and scalability of the SDSM method. 
%


\subsection{Experimental Setting} \label{sec:experiments:setting}

\begin{table}[t]
\caption{Dataset description.}
\label{tab:dataset_description}
\def\arraystretch{1.2}%
\begin{tabular}{|l|c|c|c|c|} 
\hline
& \textbf{Dataset} & \textbf{Label ($+$)} & \textbf{Label ($-$)} & \textbf{ \# Traj} ($+$, $-$)\\ 
\hline
\hline
$D_1$ & Hurricane  & Weak & Strong &  639 (75\%, 25\%)\\ 
\hline
$D_2$ & Vehicle &  Bus & Truck & 381 (28\%, 72\%)\\ 
\hline
$D_3$ & Car & Peak & Off-peak & 1,000,000 (41\%, 59\%)\\ 
\hline
\end{tabular}
\end{table}

We used three real-world trajectory datasets described in Table \ref{tab:dataset_description}.
%
Each dataset contains $n$ trajectories represented as sequences of 2D coordinates and group labels.
The first two datasets are small-size benchmark datasets which were previously studied in \cite{patel2013incorporating} and \cite{ferrero2018movelets}.
The third dataset is a large-scale dataset provided from a car insurance company
%
Dataset $D_1$ contains Atlantic hurricane trajectories between 1950 and 2008\footnote{http://weather.unisys.com/hurricane/atlantic/}. 
Each trajectory was classified by Saffir-Simpson scale from 0 to 5 (Scale 0 for the weakest and scale 5 for the strongest). 
We created two groups of hurricanes: weak (Scale 0, 1 and 2) and strong (Scale 3, 4, 5). 
%
%
The number of trajectories for each group is 480 (12,544 points) and 159 (7,289 points), respectively.
Dataset $D_2$ contains trajectories of bus and truck which were collected around Athens metropolitan area in Greece\footnote{http://chorochronos.datastories.org/}. 
The number of trajectories is 108 (66,096 points) and 273 (112,203 points) for bus group and truck group, respectively.
Dataset $D_3$ contains 1,000,000 trajectories of cars during the period between April 2018 and June 2018 provided by a car insurance company, in which each trajectory was pre-processed so that any disclosure of personal identification is avoided. 
We divided the dataset into two groups: Peak and Off-peak. 
Peak group includes trajectories whose timestamp are between 7:00 and 9:00 or 17:00 and 19:00 and the rest of trajectories were labeled as Off-peak. 
The number of trajectories is 407,879 (65,100,455 points) and 592,122 (93,001,502 points) for Peak group and Off-peak group, respectively.

\begin{table}[t]
\caption{Parameter setting.}
\label{tab:parameter_setting}
\begin{tabular}{|l|C{1.4cm}|C{1.2cm}|C{1.2cm}|}
\hline
 & \textbf{Hurricane} & \textbf{Vehicle} & \textbf{Car} \\
\hline
\hline
Minimum length $L$ &  [5,7] & [5,7] & 10 \\
\hline
Distance threshold $\veps$ &  1 & 20 &  4\\
\hline
$K$ (Distance function) & \multicolumn{3}{|c|}{5}\\
\hline
No. of permutations $B$ & \multicolumn{3}{|c|}{1000}\\
\hline
Significance level $\alpha$ & \multicolumn{3}{|c|}{0.05}\\

\hline
\end{tabular}
\end{table}

The list of tuning parameters for each dataset are summarized in Table \ref{tab:parameter_setting}. 
Here, we set the minimum length $L$ so that the extracted sub-trajectories are practically meaningful and interesting for the analysts.
The tuning parameter $K$ was set as $K=5$ for all the datasets since $K$ must be no greater than $L$.
For multiple testing procedure, we generated $B = 1000$ permuted datasets and the FWER was controlled at significance level $\alpha = 0.05$. 
We used two-sided FET for testing the statistical significance of the discriminative ability of each trajectory.

We used Hurricane and Vehicle datasets as benchmark datasets for investigating the effect of the tuning parameter $L$ on the extracted sub-trajectories and the computation cost.
The large-scale car dataset was analyzed only with $L = 10$, which was determined by practical data analysis viewpoint.
For the comparison experiments with the two benchmark datasets, we executed the code on a single CPU: Intel(R) Xeon(R) CPU E5-2687W v4 @ 3.00GHz. 
For Car dataset, we used Grid Engine with 256 cores.



\subsection{Experimental Results}

\paragraph{\textbf{Effect of Minimum Length Parameter $L$}}

\begin{figure}[t]
\begin{subfigure}[t]{0.5\linewidth}
\centering
\includegraphics[width=\linewidth]{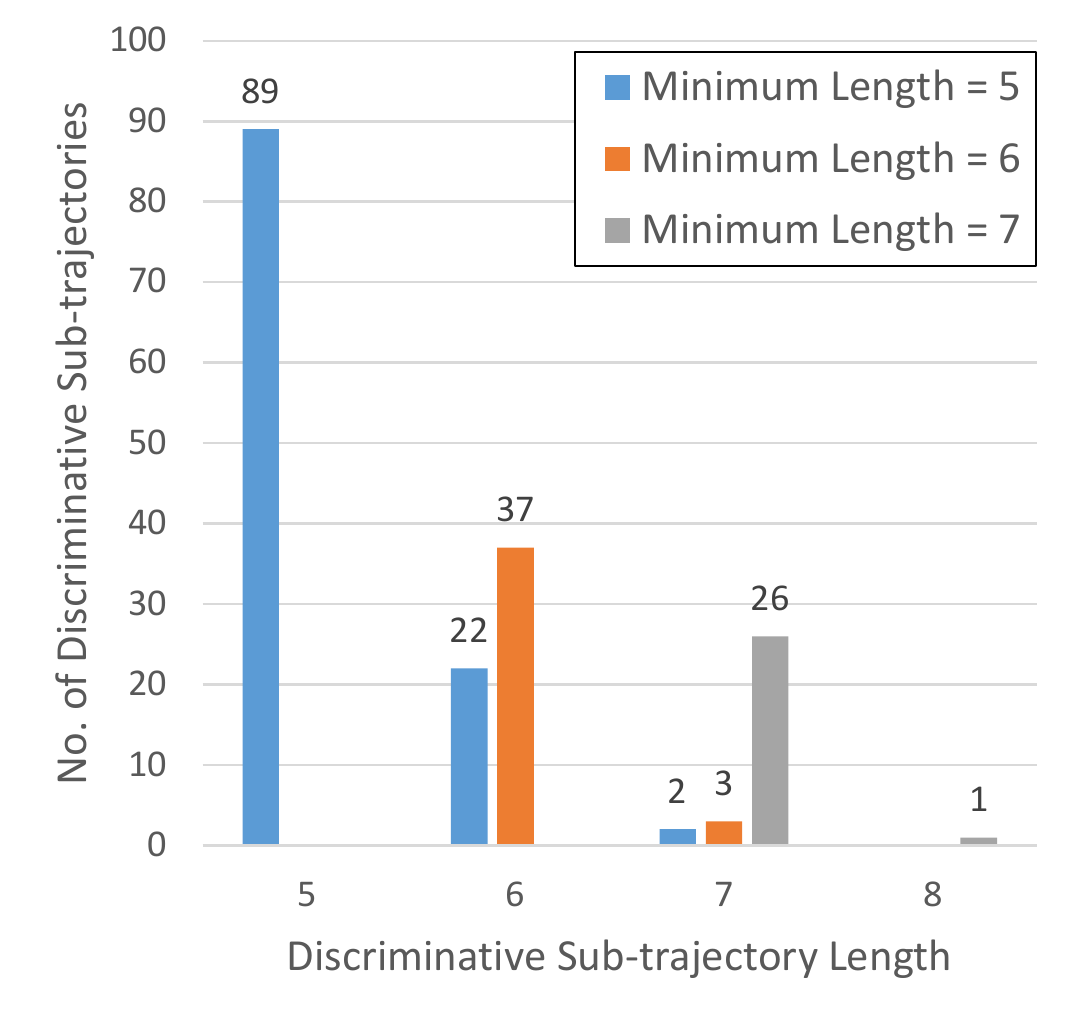}
\caption{Hurricane dataset.}
\label{fig:sub_1}
\end{subfigure}%
\begin{subfigure}[t]{0.5\linewidth}
\centering
\includegraphics[width=\linewidth]{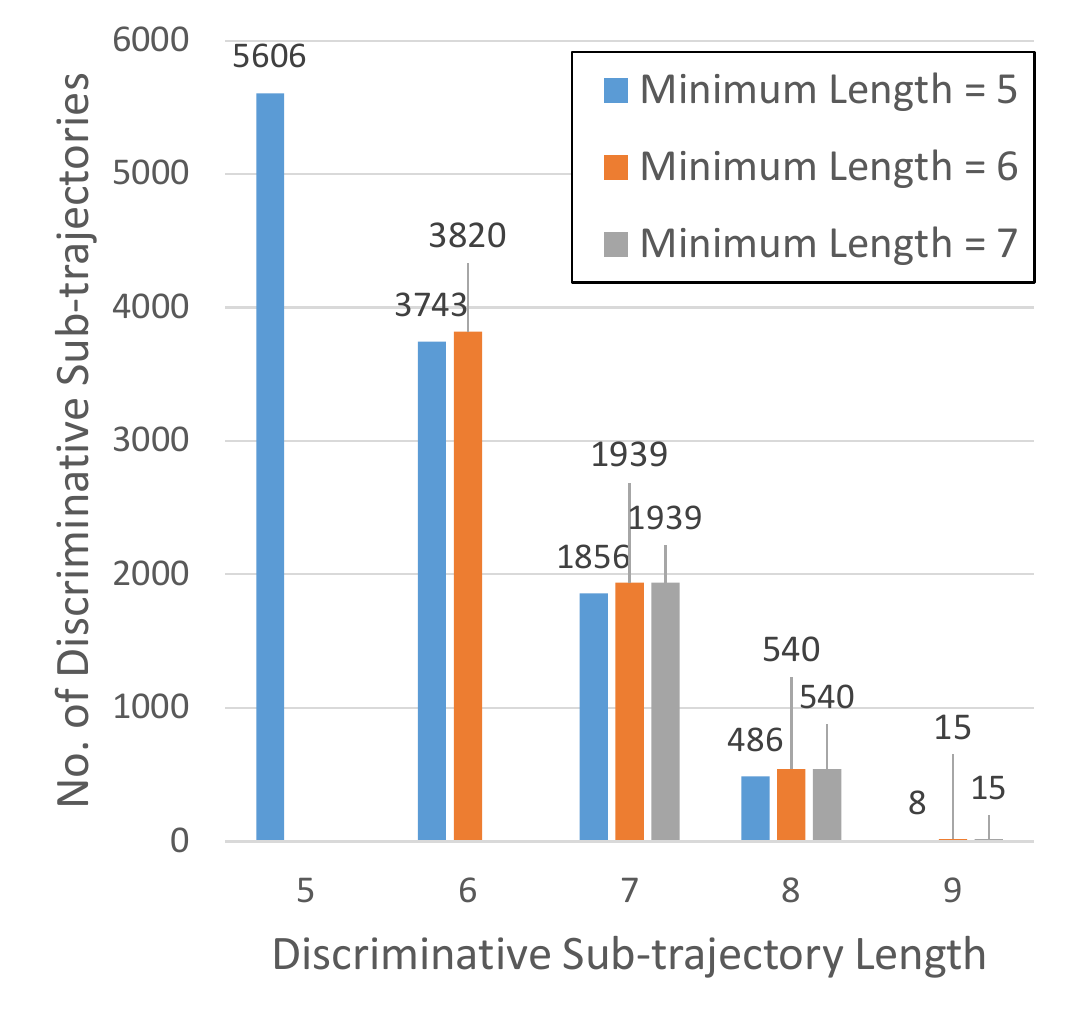}
\caption{Vehicle dataset.}
\label{fig:sub_2}
\end{subfigure}%
\caption{Discovered discriminative sub-trajectories according to different values of minimum length $L$.}
\label{fig:ex_different_min_length}
\end{figure}

Figure \ref{fig:ex_different_min_length} shows the results of discovered sub-trajectories according to different values of minimum length $L$ for Hurricane dataset  (\ref{fig:sub_1}) and Vehicle dataset (\ref{fig:sub_2}). 
In Figure \ref{fig:sub_1}, when $L=5$, the numbers of discovered sub-trajectories with length 6 and 7 were small. 
However, it increased when $L=6$.
Note that when $L=5$ or $6$, the number of sub-trajectories with length 7 was very small and there was no sub-trajectories with length 8. 
Nevertheless, when increasing $L$ to 7, the number of sub-trajectories with length 7 significantly increased and one sub-trajectory with length 8 was also found. 
In Figure \ref{fig:sub_2}, when increasing the minimum length from 5 to 7, the number of discriminative sub-trajectories of length 6, 7, 8 and 9 also increased. 
This results can be interpreted as follows. 
The number of short sub-trajectories is usually very large, and considering those sub-trajectories in multiple testing may lead to very small adjusted significance level $\delta^{\ast}$. 
Then, it would be difficult to discover statistically significant long sub-trajectories. 
On the other hand, if the minimum length $L$ is large, only longer sub-trajectories are involved in the multiple testing procedure. 
Thus, the probability of discovering longer sub-trajectories would be higher. 
As the examples indicate the determination of appropriate minimum length $L$ is very important depending on the interest of analysts.



\paragraph{\textbf{Comparison of Calculation Cost}}

\begin{table}[t]
\caption{Comparison of calculation time (sec.)}

\label{tab:comparison}

\begin{center}
\begin{tabular}{| c | c | C{2.5cm} | C{2.5cm} |}
\hline
Dataset & $L$ & \textbf{SDSM} & \textbf{Westfall-Young}\\ 
\hline
\hline
\multirow{3}{*}{Hurricane}& 5 & 1384.25 & 19007.85\\\cline{2-4}
& 6 & 494.76 & 14088.80\\\cline{2-4}
& 7 & 186.17 & 11484.77 \\
\hline
\hline
\multirow{3}{*}{Vehicle}& 5 & 3821.33 & 50815.35\\\cline{2-4}
& 6 & 2110.30 & 39944.19 \\\cline{2-4}
& 7 & 986.03 & 30019.97 \\
\hline
\end{tabular}
\end{center}
\label{tab:comparison}
\end{table}

Tables \ref{tab:comparison} shows the comparison of the computational costs between the original WY method and the SDSM method. 
We can see that the computation time decreases as the minimum length $L$ increases because the number of hypotheses decreases with the increase of $L$. 
For both Hurricane and Vehicle datasets, the SDSM method is significantly faster than the original WY method.  For the large-scale Car dataset, the original WY method could not complete the task even with the same Grid Engine in a realistic time.



\paragraph{\textbf{Results on Hurricane Data}}

\begin{table}[t]
\captionsetup{justification=centering}
\caption{\textit{p\_values} of discriminative data for Hurricane. \\$L = 5$, $\delta^{\ast} = 6.08\mathrm{e}{-5}$}
\label{tab:hurricane_discriminative_sub_trajectories}
\begin{tabular}{|c|c|c|c|} 
\hline
\textbf{sid} & \textbf{support($+$)} & \textbf{support($-$)} & \textbf{adjusted \textit{p}-value}\\ 
\hline
\hline
001 & 0 & 11 & 0.00014\\ 
\hline
002 & 0 & 11 & 0.00014\\ 
\hline
003 & 1 & 12 & 0.00034\\ 
\hline
004 & 1 & 12 & 0.00034\\ 
\hline
005 & 4 & 15 & 0.00062\\ 
\hline
006 & 4 & 14 & 0.00211\\ 
\hline
007 & 3 & 13 & 0.00212 \\ 

\hline
... & ... & ... & ...\\ 
\hline
113 & 0 & 7 & 0.04376\\ 
\hline
\end{tabular}
\end{table}

Table \ref{tab:hurricane_discriminative_sub_trajectories} shows the result of discovered sub-trajectories with $L = 5$. 
Most of them are statistically associated with strong group. 
Those discovered sub-trajectories are visualized in Figure \ref{fig:ex_hurricane}. 
Weak and strong hurricanes are respectively shown in green and pink. 
Discriminative sub-trajectories are shown in red. 
We can see that many strong hurricanes share the same movements from east to west. 
Some of them share the same movements along a curve, the direction changes from east-to-west to south-to-north.



\paragraph{\textbf{Results on Vehicle Data}}

\begin{table}[t]
\captionsetup{justification=centering}
\caption{\textit{p\_values} of discriminative data for Vehicle. \\$L = 5$, $\delta^{\ast} =  2.17\mathrm{e}{-5}$}

\label{tab:vehicle_discriminative_sub_trajectories}
\begin{tabular}{|c|c|c|c|} 
\hline
\textbf{sid} & \textbf{support($+$)} & \textbf{support($-$)} & \textbf{adjusted \textit{p}-value}\\ 
\hline
\hline
01 & 13 & 0 & 0.00011\\ 
\hline
02 & 13 & 0 & 0.00011\\ 
\hline
... & ... & ... & ...\\ 
\hline
23 & 9 & 0 &  0.02197\\ 
\hline
\hline
24 & 0 & 102 &  5.16e-15\\ 
\hline
25 & 0 & 101 &  1.05e-14\\ 
\hline
... & ... & ... & ...\\ 
\hline
11,699 &  0 & 33 &  0.04639\\ 
\hline

\end{tabular}
\end{table}

\begin{figure}[t]
\captionsetup{justification=centering}
\centering
\includegraphics[width=0.95\linewidth]{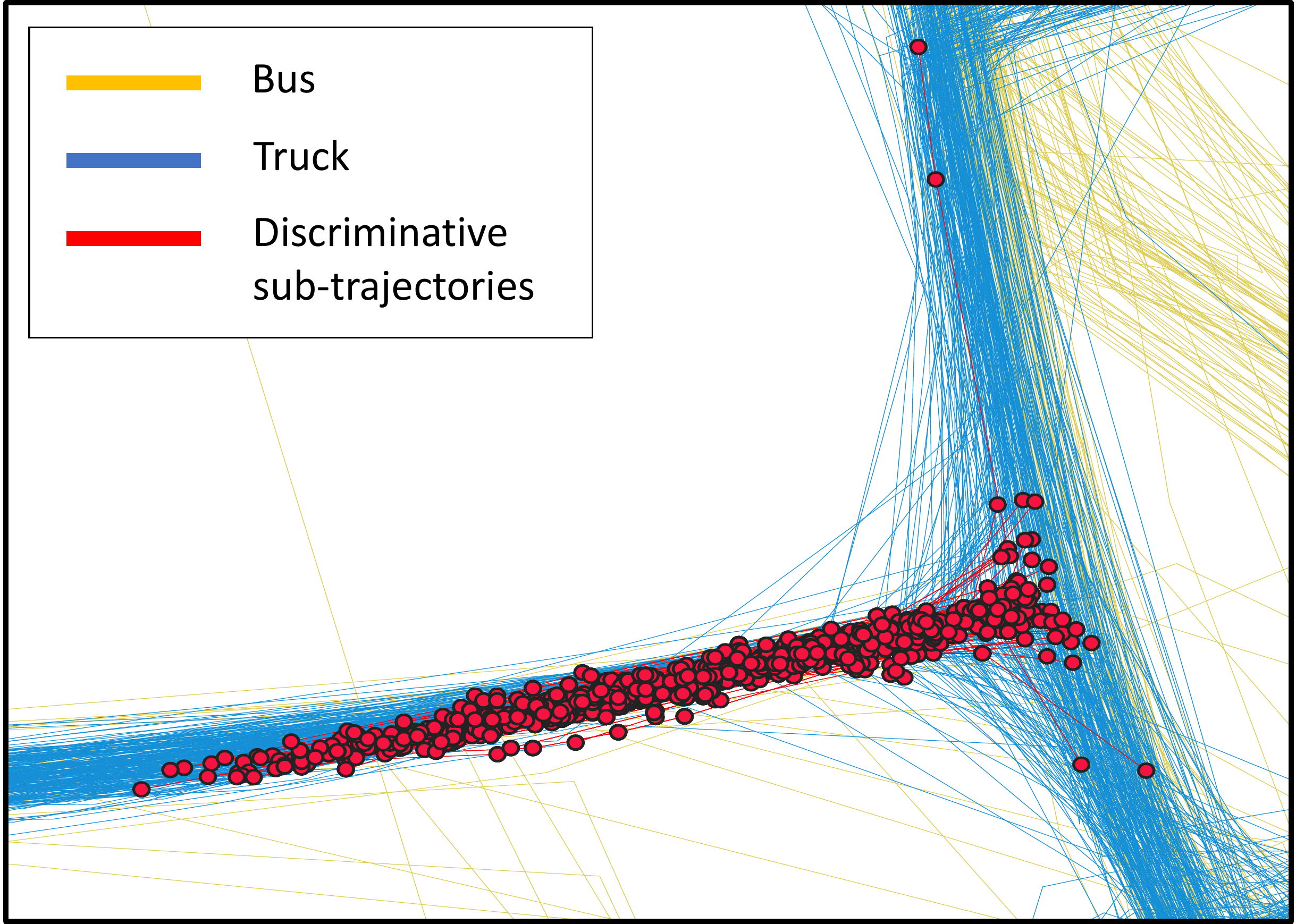}
\caption{Discriminative sub-trajectories of Truck group in vehicle dataset.}
\label{fig:ex_vehicle}
\end{figure}

Table \ref{tab:vehicle_discriminative_sub_trajectories} shows the results on Vehicle dataset with $L = 5$. 
The number of discovered discriminative sub-trajectories is 11,699. 
The first half of the table shows sub-trajectories that are associated with Bus group and the second half shows those associated with Truck group. 
The discriminative sub-trajectories of Truck group are shown in Figure \ref{fig:ex_vehicle}. 
Trajectories of Bus group and Truck group are shown in yellow and blue, respectively. 
Discriminative sub-trajectories are shown in red.



\paragraph{\textbf{Results on Car Data}}

\begin{figure}[t]
\captionsetup{justification=centering}
\centering
\includegraphics[width=0.935\linewidth]{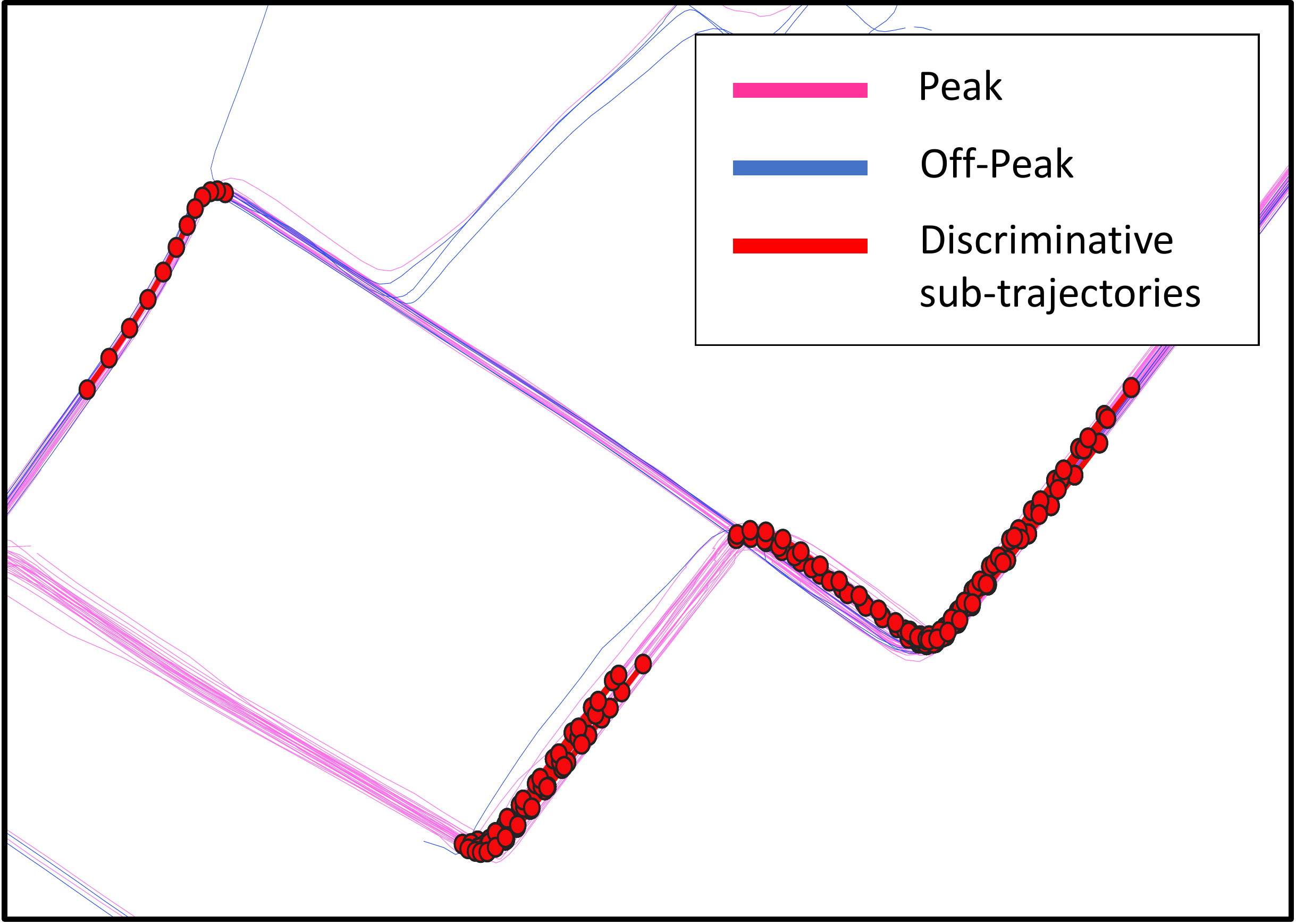}
\caption{Discriminative sub-trajectories of Peak group in car dataset.}
\label{fig:ex_car_1}
\end{figure}

\begin{figure}[t]
\captionsetup{justification=centering}
\centering
\includegraphics[width=0.935\linewidth, height=0.235\textheight]{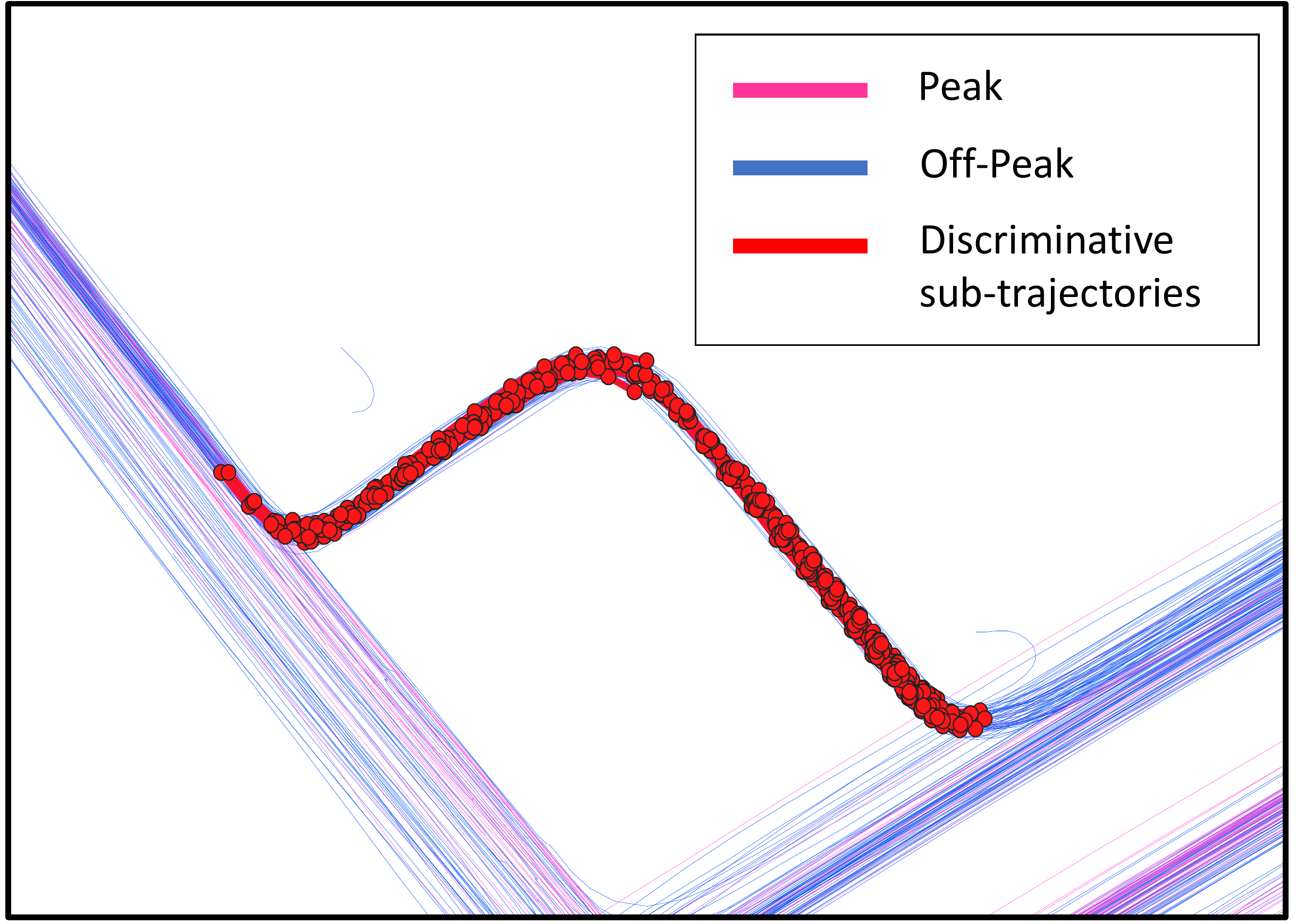}
\caption{Discriminative sub-trajectories of Off-Peak group in car dataset.}
\label{fig:ex_car_2}
\end{figure}

Table \ref{tab:car_discriminative_sub_trajectories} shows the results on Car dataset. 
The total number of discovered discriminative sub-trajectories is 848,469. 
The results for Peak and Off-peak  are respectively shown in the first half and second half of the table. 
The discriminative sub-trajectories for each group are shown in Figures \ref{fig:ex_car_1} and \ref{fig:ex_car_2}. 
We use pink for representing trajectories of Peak group and blue for those in Off-peak group. 
Discriminative sub-trajectories are shown in red. 
Figure \ref{fig:ex_car_1} shows places where many cars pass in Peak rather than Off-peak, Figure \ref{fig:ex_car_2} shows places where many cars pass in Off-peak. 
Based on these results, we might be able to recommend appropriate paths to avoid congestion according to peak or off-peak hours.

\begin{table}[t]
\captionsetup{justification=centering}
\caption{\textit{p\_values} of discriminative data for Car. \\$L = 10$, $\delta^{\ast} =  4.35\mathrm{e}{-8}$}

\label{tab:car_discriminative_sub_trajectories}
\begin{tabular}{|c|c|c|c|} 
\hline
\textbf{sid} & \textbf{support($+$)} & \textbf{support($-$)} & \textbf{adjusted \textit{p}-value}\\ 
\hline
\hline
01 & 82 & 1 & 6.56e-25\\ 
\hline
02 & 77 & 0 & 1.17e-24\\ 
\hline
... & ... & ... & ...\\ 
\hline
708,139 & 51 & 18 &   0.04972\\ 
\hline
\hline
708,140 & 0 & 75 &  1.23e-11\\ 
\hline
708,141 & 0 & 73 &  4.02e-11\\ 
\hline
... & ... & ... & ...\\ 
\hline
848,469 &  2 & 43 &  0.04999\\ 
\hline

\end{tabular}
\end{table}

%% file: sec5.tex
\section{Conclusions} \label{sec:conclusions}

In this paper, we have introduced a novel method, called SDSM, for mining discriminative patterns of moving objects, which are represented in the form of sub-trajectories. 
We performed experiments on three real-world datasets: hurricane, vehicle and car to show the effectiveness and the scalability of the SDSM method. 
In conclusion, we believe that this paper provides a new paradigm for statistically discriminative moving pattern mining. 
It helps data analysts to discover specific moving patterns while controlling the risk of false discoveries.